\documentclass[runningheads]{llncs}
\usepackage{graphicx}
\usepackage{comment}
\usepackage{amsmath,amssymb} 
\usepackage{color}
\usepackage{threeparttable}
\usepackage{multirow}
\usepackage{multicol}
\usepackage{setspace}
\usepackage{adjustbox}

\begin{document}
	\pagestyle{headings}
	\mainmatter
	\def\ECCVSubNumber{1086}  
	
	\title{MuCAN: Multi-Correspondence Aggregation Network for Video Super-Resolution}

	\titlerunning{MuCAN: Multi-Correspondence Aggregation Network}
	%
\author{Wenbo Li\inst{1} \and Xin Tao\inst{2} \and Taian Guo\inst{3} \and Lu Qi\inst{1} \and Jiangbo Lu\inst{4} \and Jiaya Jia\inst{1,4}}
	\authorrunning{W. Li, X. Tao, T. Guo, L. Qi, J. Lu, J. Jia}
	%
	\institute{$^{1}$The Chinese University of Hong Kong \\ $^{2}$Kuaishou Technology \ $^{3}$Tsinghua University \ ${^4}$SmartMore\\
		\email{\{wenboli,luqi,leojia\}@cse.cuhk.edu.hk} \quad \email{jiangsutx@gmail.com} \quad \email{gta17@mails.tsinghua.edu.cn} \quad \email{jiangbo@smartmore.com} }
	\maketitle
	
	\begin{abstract}
		Video super-resolution (VSR) aims to utilize multiple low-resolution frames to generate a high-resolution prediction for each frame. In this process, inter- and intra-frames are the key sources for exploiting temporal and spatial information. However, there are a couple of limitations for existing VSR methods. First, optical flow is often used to establish temporal correspondence. But flow estimation itself is error-prone and affects recovery results. Second, similar patterns existing in natural images are rarely exploited for the VSR task. Motivated by these findings, we propose a temporal multi-correspondence aggregation strategy to leverage similar patches across frames, and a cross-scale nonlocal-correspondence aggregation scheme to explore self-similarity of images across scales. Based on these two new modules, we build an effective multi-correspondence aggregation network (MuCAN) for VSR. Our method achieves state-of-the-art results on multiple benchmark datasets. Extensive experiments justify the effectiveness of our method.
	\end{abstract}

	\section{Introduction}\label{sec:intro}
	
	Super-resolution (SR) is a fundamental task in image processing and computer vision, which aims to reconstruct high-resolution (HR) images from low-resolution (LR) ones. While single-image super-resolution methods are mostly based on spatial information, video super-resolution (VSR) needs to exploit temporal structure from multiple neighboring frames to recover missing details. VSR can be widely applied to video surveillance, satellite imagery, etc.
	
	Early methods~\cite{liu2013bayesian,ma2015handling} for VSR used various delicate image models, which are solved via optimization techniques. Recent deep-neural-network-based VSR methods \cite{kim20183dsrnet,li2019fast,tian2018tdan,wang2019edvr,tao2017detail,caballero2017real,liao2015video,ma2015handling,haris2019recurrent,sajjadi2018frame} further push the limit and set new state-of-the-arts. 
	
	In contrast to previous methods that model VSR as separate alignment and regression stages, we view this problem as an inter- and intra-frame correspondence aggregation task. Based on the fact that consecutive frames share similar content, and different locations within a single frame may contain similar structures (known as self-similarity~\cite{kindermann2005deblurring,protter2008generalizing,glasner2009super}), we propose to aggregate similar content from multiple corresponding patches to better restore HR results.
	
\begin{figure}[t]
		\begin{center}
			\includegraphics[width=0.8\linewidth]{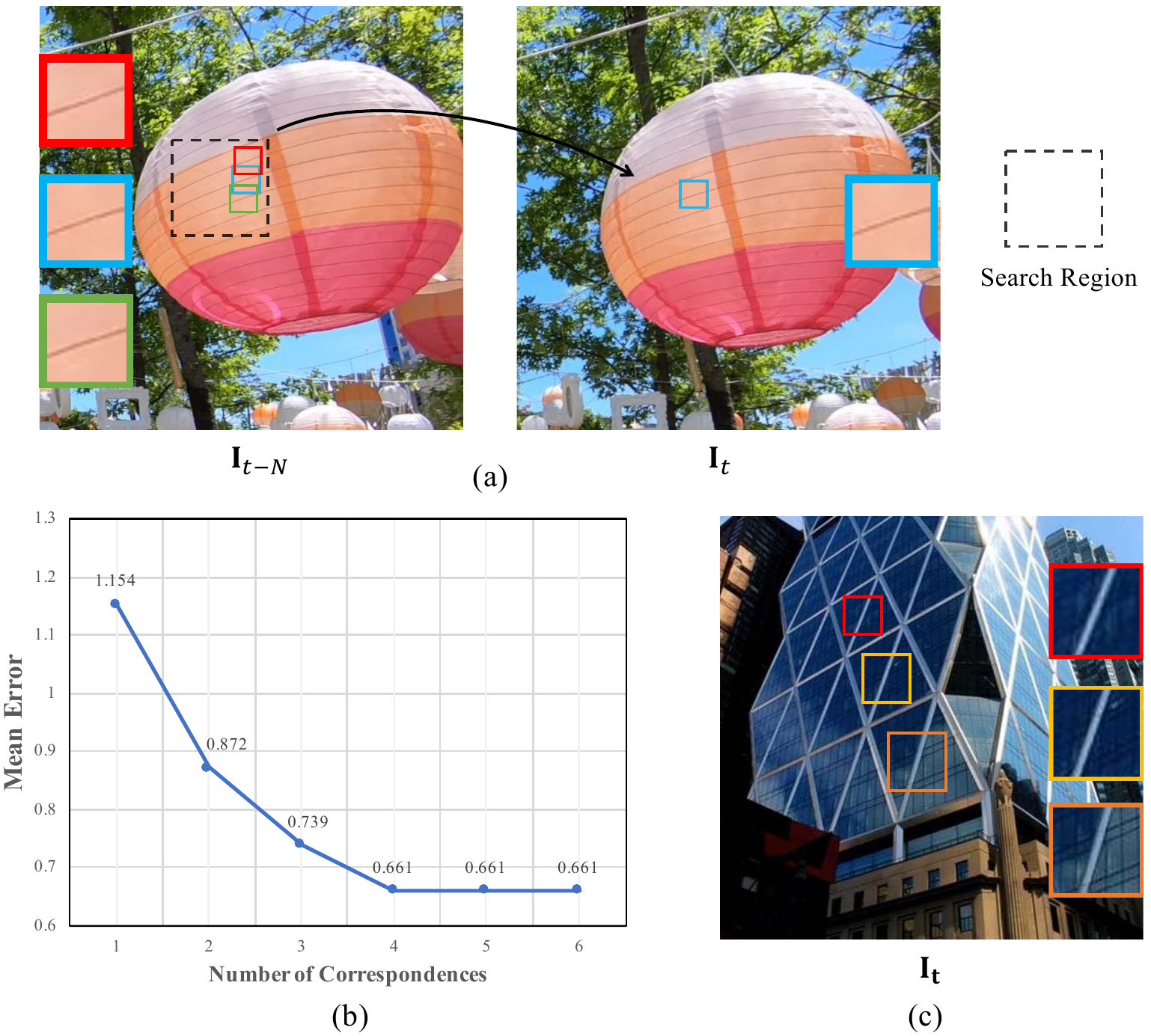}
		\end{center}
		\vspace{-0.2in}
		\caption{Inter- and intra-frame correspondence. (a) Inter-frame correspondence estimated from temporal frames (e.g., ${\textbf I}_{t-N}$ at time $t-N$ and  $\textbf{I}_{t}$ at time $t$) can be leveraged for VSR. For a patch in ${\textbf I}_{t}$, there are actually multiple similar patterns within a co-located search region in ${\textbf I}_{t-N}$. (b) Mean error of optical flow estimated by different numbers of inter-frame correspondences on a subset of MPI Sintel Flow dataset. (c) Similar patterns existing over different scales in an image ${\textbf I}_t$. }
		\label{fig:corr}
		\vspace{-0.15in}
	\end{figure}

	\paragraph{Inter-frame Correspondence} Motion compensation (or alignment) is usually an essential component for most video tasks to handle displacement among frames. A majority of methods~\cite{caballero2017real,tao2017detail,sajjadi2018frame} design specific sub-networks for optical flow estimation. In~\cite{kim20183dsrnet,li2019fast,jo2018deep,huang2015bidirectional}, motion is implicitly handled using Conv3D or recurrent networks. Recent methods~\cite{tian2018tdan,wang2019edvr} utilize deformable convolution layers~\cite{dai2017deformable} to explicitly align feature maps using learnable offsets. All the methods establish explicit or implicit one-on-one pixel correspondence between frames. However, motion estimation may suffer from inevitable errors and there is no chance for wrongly estimated mapping to locate correct pixels. Thus, we resort to another line of solutions to simultaneously consider multiple correspondence candidates for each pixel, as illustrated in Figure~\ref{fig:corr}(a).  
	
	In order to preliminarily validate our idea, we estimate optical flow with a simple patch-matching strategy on the MPI Sintel Flow dataset. After obtaining top-$K$ most similar patches as correspondence candidates, we calculate the Euclidean distance between the best-performing one and ground-truth flow. As shown in Figure~\ref{fig:corr}(b), it is clear that the better result is obtained by taking into consideration more correspondences for a pixel. Inspired by this finding, we propose a {\it temporal multi-correspondence aggregation module} (TM-CAM) for alignment. It uses top-$K$ most similar feature patches as supplement. More specifically, we design a pixel-adaptive aggregation strategy, which will be detailed in Sec.~\ref{sec:method}. Our module is lightweighted and, more interestingly, can be easily integrated into many common frameworks. It is surprisingly robust to visual artifacts, as demonstrated in Sec.~\ref{sec:exp}.
	
	\paragraph{Intra-frame Correspondence} From another perspective, similar patterns within each frame as shown in Figure~\ref{fig:corr}(c) can also benefit detail restoration, which has been verified in several previous low-level tasks~\cite{kindermann2005deblurring,protter2008generalizing,glasner2009super,yang2010exploiting,zhang2010bregmanized,huang2015single}. This line is still new for VSR. For existing methods in VSR, the common way to explore intra-frame information is to introduce a U-net-like~\cite{ronneberger2015u} or deep structure, so that a large but still local receptive field is covered. We notice that valuable information may not always come from neighboring positions. In fact, similar patches within nonlocal locations or across scales may also be beneficial.
	
	Accordingly, we in this paper design a new {\it cross-scale nonlocal-correspondence aggregation module} (CN-CAM) to exploit the multi-scale self-similarity property of natural images. It aggregates similar features across different levels to recover more details. The effectiveness of this module is verified in Sec.~\ref{sec:exp}. 
	
	The contribution of this paper is threefold.
	\begin{itemize}
		\item We design a multi-correspondence aggregation network (MuCAN) to deal with video super-resolution in an end-to-end manner. It achieves state-of-the-art performance on multiple benchmark datasets.
		\item Two effective modules are proposed to make decent use of temporal and spatial information. The temporal multi-correspondence aggregation module (TM-CAM) conducts motion compensation in a robust way. The cross-scale nonlocal-correspondence aggregation module (CN-CAM) explores similar features from multiple spatial scales.
		\item We introduce an edge-aware loss that enables the proposed network to generate better refined edges.
	\end{itemize}
	
	
	\section{Related Work}
	
	Super-resolution is a classical task in computer vision. Early work used example-based~\cite{freeman2002example,glasner2009super,yang2010image,freedman2011image,timofte2013anchored,timofte2014a+,schulter2015fast}, dictionary learning~\cite{yang2012coupled,perez2016psyco} and self-similarity~\cite{yang2010exploiting,huang2015single} methods. Recently, with the rapid development of deep learning, super-resolution reached a new level. In this section, we briefly discuss deep learning based approaches from two lines, i.e., single-image super-resolution (SISR) and video super-resolution (VSR).
	
	\subsection{Single-Image Super-Resolution}
	SRCNN~\cite{dong2014learning} is the first method that employs a deep convolutional neural network in the image super-resolution task. It has inspired several following methods~\cite{dong2016accelerating,kim2016accurate,shi2016real,kim2016deeply,lai2017deep,tai2017image,haris2018deep,zhang2018image,zhang2018residual}. For example, Kim {\it et~al}.~\cite{kim2016accurate} proposed a residual learning strategy using a 20-layer depth network, which yielded significant accuracy improvement. Instead of applying commonly used bicubic interpolation, Shi {\it et~al}.~\cite{shi2016real} designed a sub-pixel convolution network to effectively upsample low-resolution input. This operation reduces computational complexity and enables a real-time network. Taking advantage of high-quality large image datasets, more networks, such as DBPN~\cite{haris2018deep}, RCAN~\cite{zhang2018image}, and RDN~\cite{zhang2018residual}, further improve the performance of SISR.
	
	\subsection{Video Super-Resolution}
	
	Video super-resolution takes multiple frames into consideration. Based on the way to aggregate temporal information, previous methods can be roughly grouped into three categories.
	
	The first group of methods process video sequences without any explicit alignment. For example, methods of~\cite{kim20183dsrnet,li2019fast} utilize 3D convolutions to directly extract features from multiple frames. Although this approach is simple, the computational cost is typically high. Jo {\it et~al}.~\cite{jo2018deep} proposed dynamic upsampling filters to avoid explicit motion compensation. However, it stands the chance of ignoring informative details of neighboring frames. Noise in the misaligned regions can also be harmful.
	
	The second line~\cite{liao2015video,ma2015handling,kappeler2016video,caballero2017real,liu2017robust,tao2017detail,sajjadi2018frame,haris2019recurrent} is to use optical flow to compensate motion between frames. Methods of~\cite{liao2015video,kappeler2016video} first obtain optical flow using classical algorithms and then build a network for high-resolution image reconstruction. Caballero {\it et~al}. \cite{caballero2017real} integrated these two steps into a single framework and trained it in an end-to-end way. Tao {\it et~al}.~\cite{tao2017detail} further proposed sub-pixel motion compensation to reveal more details. No matter if optical flow is predicted independently, this category of methods needs to handle two relatively separated tasks. Besides, the estimated optical flow critically affects the quality of reconstruction. Because optical flow itself is a challenging task especially for large-motion scenes, the resulting accuracy cannot be guaranteed.
	
	The last line~\cite{tian2018tdan,wang2019edvr} conducts deformable convolution networks~\cite{dai2017deformable} to accomplish video super-resolution. For example, EDVR proposed in \cite{wang2019edvr} extracts and aligns features at multiple levels, and achieves reasonable performance. The deformable network is however sensitive to the input patterns, and may give rise to noticeable reconstruction artifacts due to unreasonable offsets.
	
	
	\section{Our Method}\label{sec:method}
	
	\begin{figure}[t]
		\begin{center}
			\includegraphics[width=1.0\linewidth]{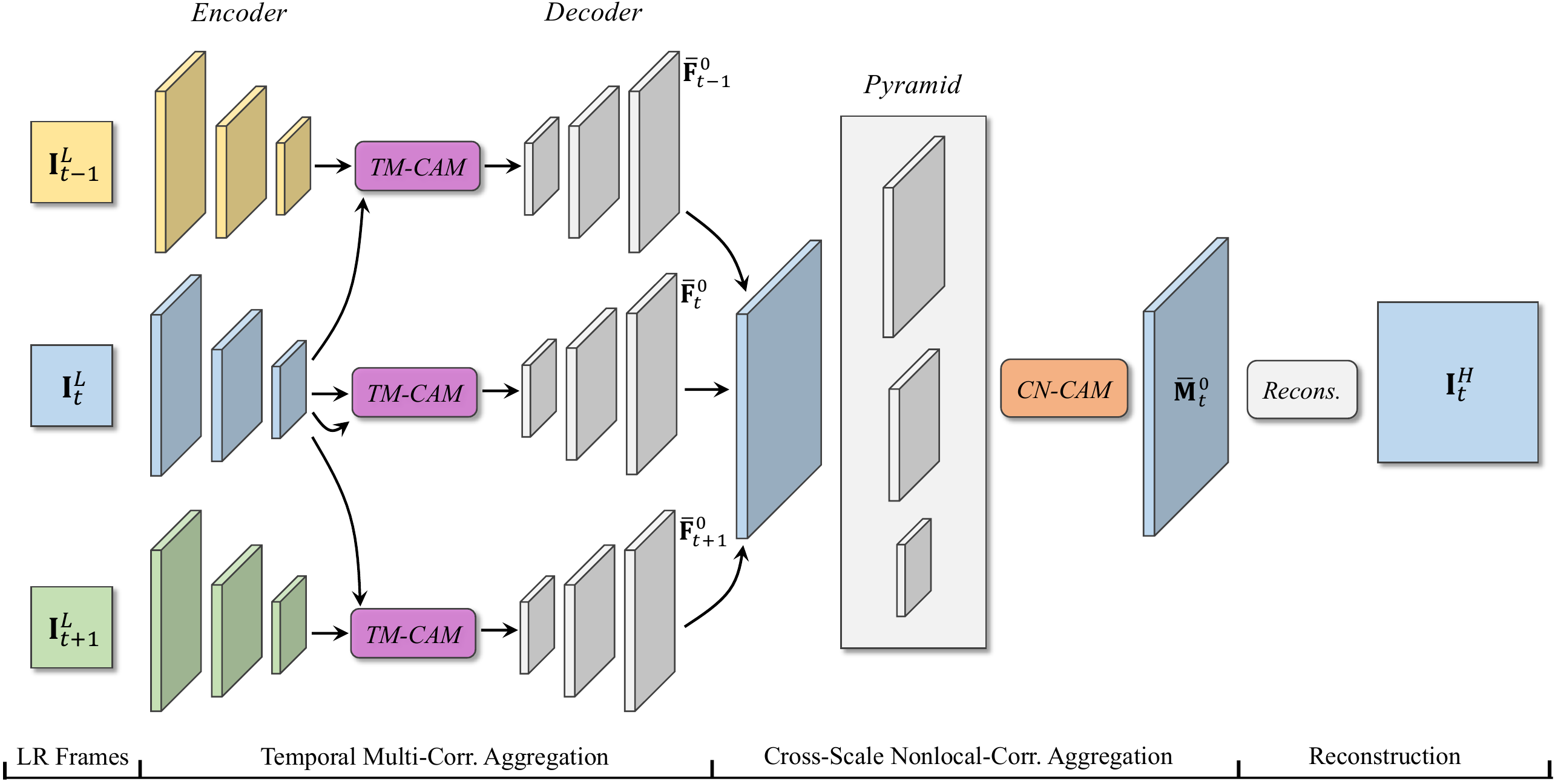}
		\end{center}\vspace{-0.15in}
		\caption{Architecture of our multi-correspondence aggregation network (MuCAN). It contains two novel modules: temporal multi-correspondence aggregation module~(TM-CAM) and cross-scale nonlocal-correspondence aggregation module~(CN-CAM).}
		\label{fig:framework}
		\vspace{-0.1in}
	\end{figure}
	
	
	The architecture of our proposed multi-correspondence aggregation network (MuCAN) is illustrated in Figure~\ref{fig:framework}. Given $2N+1$ consecutive low-resolution frames $\{{\textbf I}^{L}_{t-N}, \dots, {\textbf I}^{L}_{t}, \dots, {\textbf I}^{L}_{t+N}\}$, our framework predicts a high-resolution central image ${\textbf I}^{H}_{t}$.
	It is an end-to-end network consisting of three modules: a temporal multi-correspondence aggregation module (TM-CAM), a cross-scale nonlocal-correspondence aggregation module (CN-CAM), and a reconstruction module. The details of each module are given in the following subsections.

	\subsection{Temporal Multi-Correspondence Aggregation Module}
	\label{part:tm-cam}
	
	\begin{figure}[t]
		\begin{center}
			\includegraphics[width=0.88\linewidth]{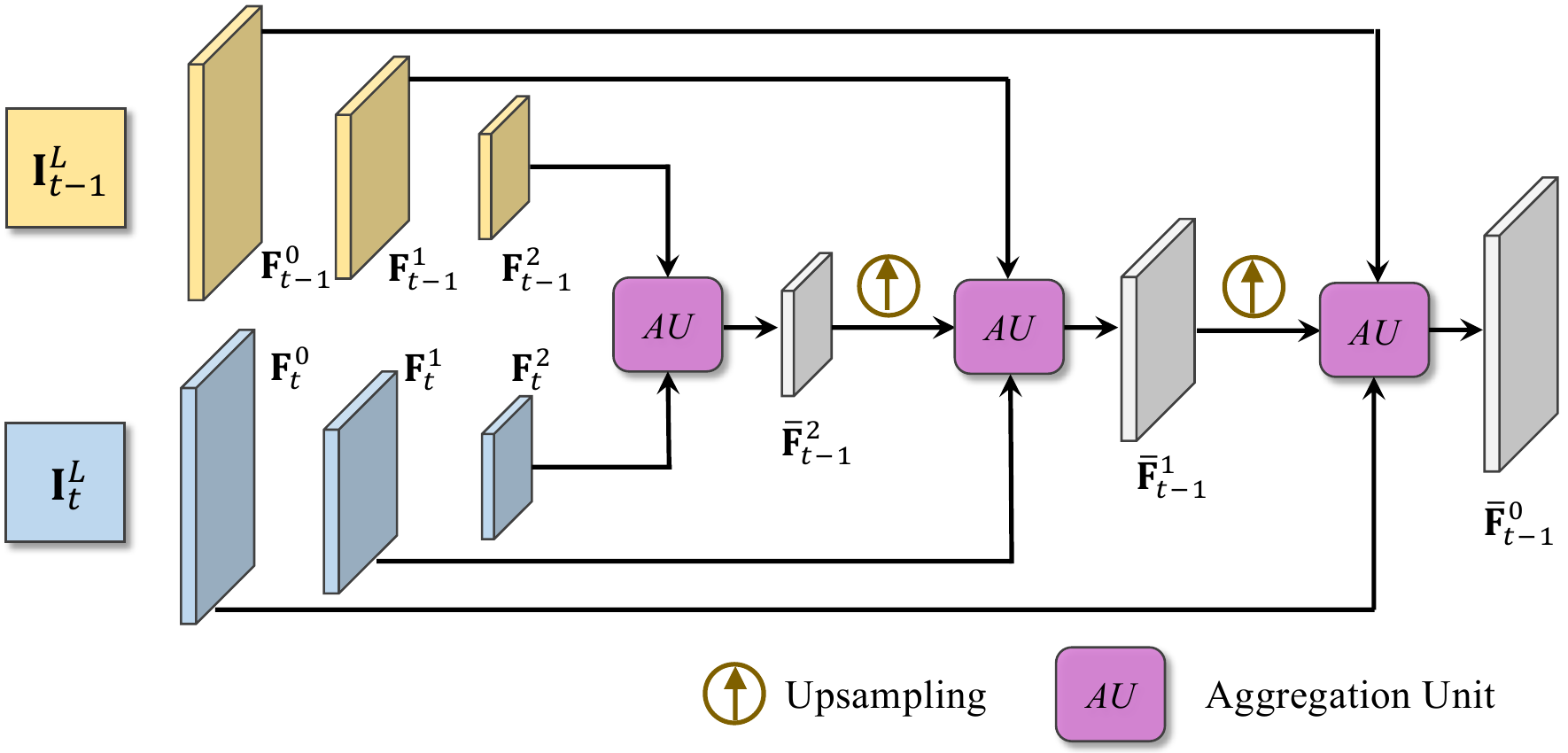}
		\end{center}\vspace{-0.2in}
		\caption{Structure of temporal multi-correspondence aggregation module~(TM-CAM).}
		\label{fig:tm-cam}
	\end{figure}
	
	\begin{figure}[h]
		\begin{center}
			\includegraphics[width=0.95\linewidth]{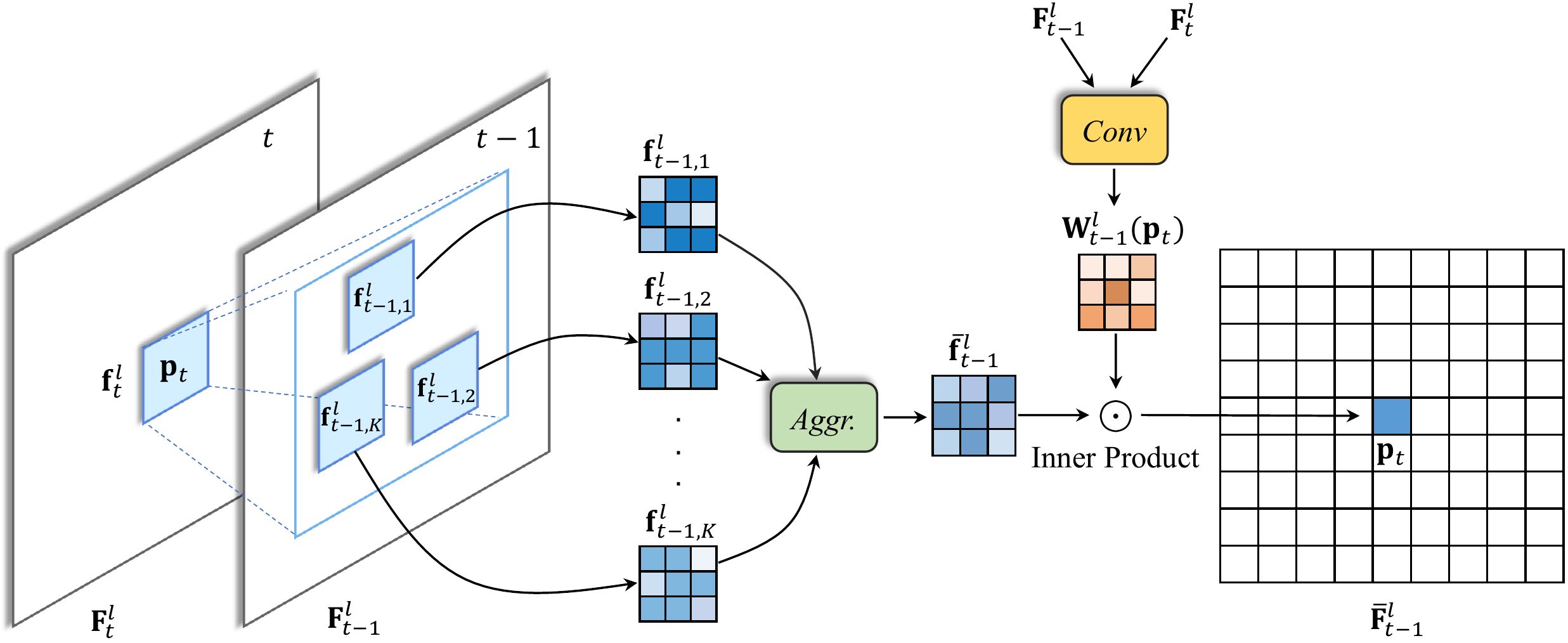}
		\end{center}\vspace{-0.2in}
		\caption{Aggregation unit in TM-CAM. It aggregates multiple inter-frame correspondences to recover a given pixel at ${\textbf p}_{t}$.}
		\label{fig:au}
		\vspace{-0.15in}
	\end{figure}
	
	Camera or object motion between neighboring frames has its pros and cons. On the one hand, large motion needs to be eliminated to build correspondence among similar content. On the other hand, accuracy of small motion (at sub-pixel level) is very important, which is the source to draw details. Inspired by the work of \cite{ranjan2017optical,sun2018pwc}, we design a hierarchical correspondence aggregation strategy to handle large and subtle motion simultaneously. 
	
	As shown in Figure~\ref{fig:tm-cam}, given two neighboring LR images ${\textbf I}^{L}_{t-1}$ and ${\textbf I}^{L}_{t}$, we first encode them into lower resolutions (level $l=0$ to $l=2$). Then, the aggregation starts in the high-level (low-resolution) stage (i.e., from $\overline{{\textbf F}}^{l = 2}_{t-1}$) compensating large motion, while progressively moving up to low-level/high-resolution stages (i.e., to $\overline{{\textbf F}}^{l = 0}_{t-1}$) for subtle sub-pixel shift. Different from many methods~\cite{caballero2017real,tao2017detail,sajjadi2018frame} that directly regress flow fields in image space, our module functions in the feature space. It is more stable and robust to noise~\cite{sun2018pwc}.
	
	The aggregation unit in Figure~\ref{fig:tm-cam} is detailed in Figure~\ref{fig:au}. A patch-based matching strategy is used since it naturally contains structural information. As aforementioned in Figure~\ref{fig:corr}(b), one-to-one mapping may not be able to capture true correspondence between frames. We thus aggregate multiple candidates to obtain sufficient context information as in Figure~\ref{fig:au}. 
	
	In details, we first locally select top-$K$ most similar feature patches, and then utilize a pixel-adaptive aggregation scheme to fuse them into one pixel to avoid boundary problems. Taking aligning ${\textbf F}^{l}_{t-1}$ to ${\textbf F}^{l}_{t}$ as an example, given an image patch ${\textbf f}^{\, l}_{t}$ (represented as a feature vector) in ${\textbf F}^{l}_{t}$, we first find its nearest neighbors on ${\textbf F}^{l}_{t-1}$. For efficiency, we define a local search area satisfying $\left\vert {\textbf p} _{t} - {\textbf p}_{t-1} \right\vert \le {\textbf d}$, where ${\textbf p}_{t}$ is the position vector of ${\textbf f}^{\, l}_{t}$. ${\textbf d}$ means the maximum displacement. We use correlation as a distance measure as in Flownet~\cite{dosovitskiy2015flownet}. For ${\textbf f}^{\, l}_{t-1}$ and ${\textbf f}^{\, l}_{t}$, their correlation is computed as the normalized inner product of
	\begin{equation}
	\begin{aligned}
	corr({\textbf f}^{\, l}_{t-1}, {\textbf f}^{\, l}_{t}) = \frac{{\textbf f}^{\, l}_{t-1}}{|| {\textbf f}^{\, l}_{t-1} ||} \cdot \frac{{\textbf f}^{\, l}_{t}}{|| {\textbf f}^{\, l}_{t} ||}\,.
	\end{aligned}
	\end{equation}
	After calculating correlations, we select top-$K$ most correlated patches (i.e., ${\bar {\textbf f}}^{\, l}_{t-1, 1} , {\bar {\textbf f}}^{\, l}_{t-1, 2} , \dots , {\bar {\textbf f}}^{\, l}_{t-1, K}$) in a descending order from ${\textbf F}^{l}_{t-1}$, and concatenate and aggregate them as
	\begin{equation}
	\begin{aligned}
	{\bar {\textbf f}}^{\, l}_{t-1} = Aggr\left( \left[ {\bar {\textbf f}}^{\, l}_{t-1, 1} , \; {\bar {\textbf f}}^{\, l}_{t-1, 2} , \; \dots , \; {\bar {\textbf f}}^{\, l}_{t-1, K} \right] \right) \,,
	\end{aligned}
	\end{equation}
	where $Aggr$ is implemented as convolution layers. Instead of assigning equal weights (e.g., $\frac{1}{9}$ when the patch size is 3), we design a pixel-adaptive aggregation strategy to enable varying aggregation patterns in different locations. The weight map is obtained by concatenating $ {\textbf F}^{l}_{t-1}$ and ${\textbf F}^{l}_{t}$ and going through a convolution layer, which has a size of $H \times W \times s^2$ when the patch size is $s \times s$. The adaptive weight map takes the form of
	\begin{equation}
	\begin{aligned}
	{\textbf W}^{l}_{t-1}= Conv \left( \left[ {\textbf F}^{l}_{t-1} , \; {\textbf F}^{l}_{t} \right] \right) \,.
	\end{aligned}
	\end{equation}
	As shown in Figure~\ref{fig:au}, the final value at position ${\textbf p}_{t}$ on the aligned neighboring frame $\bar{\textbf F}^{l}_{t-1}$ is obtained as
	\begin{equation}
	\begin{aligned}
	\bar{{\textbf F}}^{l}_{t-1}(\textbf {p}_{t}) = \bar{{\textbf f}}^{\, l}_{t-1} \cdot {\textbf W}^{l}_{t-1}({\textbf p}_{t}) \,.
	\end{aligned}
	\end{equation}
	After repeating the above steps for $2N$ times, we obtain a set of aligned neighboring feature maps $ \{ {\bar{\textbf F}}^{0}_{t-N}, \dots, {\bar{\textbf F}}^{0}_{t-1}, {\bar{\textbf F}}^{0}_{t+1}, \dots, {\bar{\textbf F}}^{0}_{t+N} \}$. To handle all frames at the same feature level, as shown in Figure~\ref{fig:framework}, we employ an additional TM-CAM, which performs self-aggregation with ${\textbf I}^{L}_{t}$ as the input and produces ${\bar{\textbf F}}^{0}_{t}$. Finally, all these feature maps are fused into a double-spatial-sized feature map by a convolution and depth-to-space operation (PixelShuffle), which is to keep sub-pixel details.
	
	\subsection{Cross-Scale Nonlocal-Correspondence Aggregation Module}
	\label{part:cn-cam}
	
\begin{figure}[t]
		\begin{center}
			\includegraphics[width=0.9\linewidth]{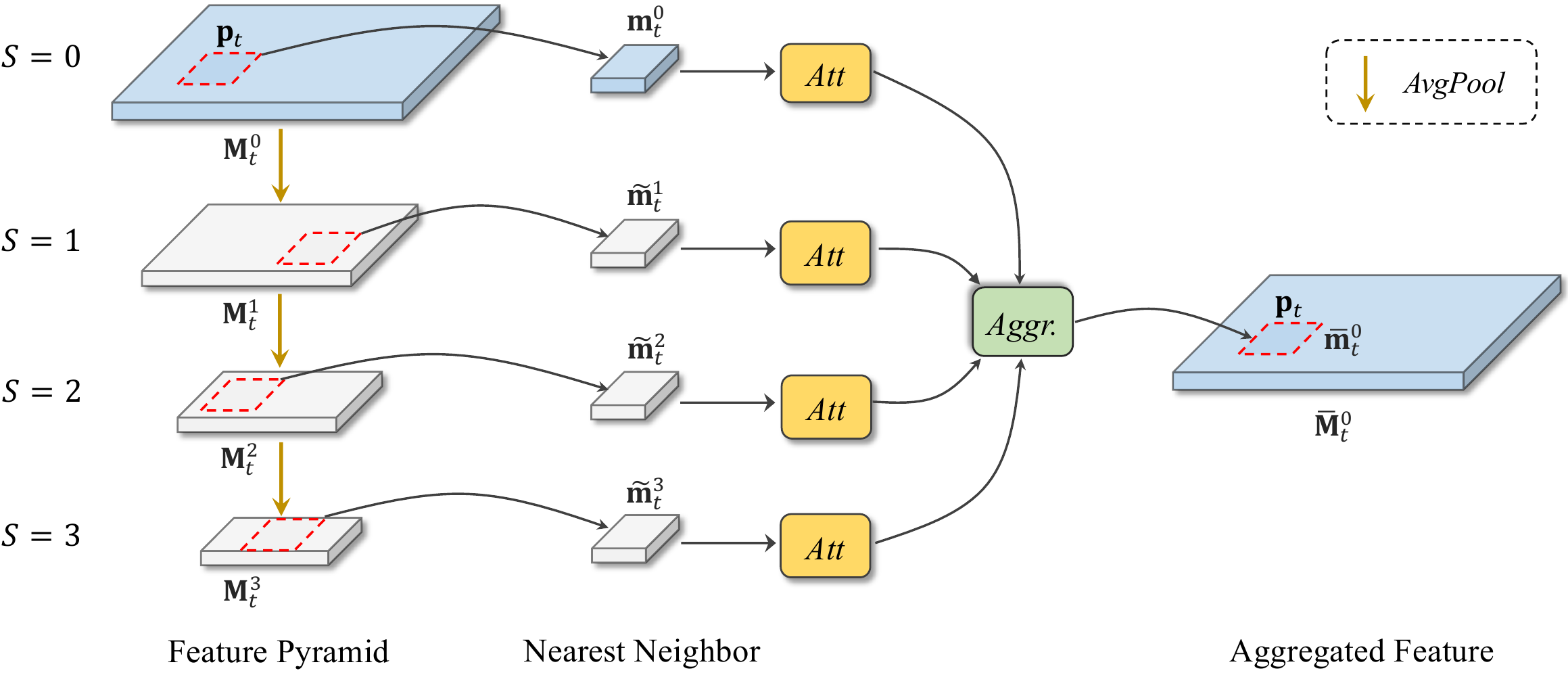}
		\end{center}\vspace{-0.15in}
		\caption{Cross-scale nonlocal-correspondence aggregation module (CN-CAM).}
		\label{fig:cn-cam}
		\vspace{-0.15in}
	\end{figure}

	Similar patterns exist widely in natural images that provide abundant texture information. Self-similarity \cite{kindermann2005deblurring,protter2008generalizing,glasner2009super,yang2010exploiting,zhang2010bregmanized} can help detail recovery. In this part, we design a cross-scale aggregation strategy to capture nonlocal correspondences across different feature resolutions, as illustrated in Figure~\ref{fig:cn-cam}.
	
	To distinguish from Sec.~\ref{part:tm-cam}, we use $\textbf{M}^{s}_{t}$ to denote feature maps at time $t$ with scale level $s$. We first downsample the input feature maps $\textbf{M}^{0}_{t}$ and obtain a feature pyramid as
	\begin{equation}
	\begin{aligned}
	{\textbf M}^{s+1}_{t} = AvgPool \left( {\textbf M}^{s}_{t} \right) , \; s=\{0, 1, 2\} \,,
	\end{aligned}
	\end{equation}
	where $AvgPool$ is the average pooling with stride 2. Given a query patch ${\textbf m}^{0}_{t}$ in $\textbf{M}^{0}_{t}$ centered at position ${\textbf p}_{t}$, we implement a non-local search on other three scales to obtain
	\begin{equation}
	\begin{aligned}
	{\tilde{\textbf m}}^{s}_{t} = NN \left( {\textbf M}^{s}_{t}, {\textbf m}^{0}_{t}\right), \; s=\{1, 2, 3\} \,,
	\end{aligned}
	\end{equation}
	where ${\tilde{\textbf m}}^{s}_{t}$ denotes the nearest neighbor (the most correlated patch) of ${\textbf m}^{0}_{t}$ in ${\textbf M}^{s}_{t}$.
	Before merging, a self-attention module~\cite{wang2019edvr} is applied to determine whether the information is useful or not. Finally, the aggregated feature $\bar{\textbf{m}}^{0}_{t}$ at position ${\textbf p}_{t}$ is calculated as
	\begin{equation}
	\begin{aligned}
	\bar{\textbf{m}}^{0}_{t}= Aggr \left(\left[ Att \left( \textbf{m}^{0}_{t} \right), Att \left( {\tilde{\textbf{m}}}^{1}_{t} \right), Att\left( {\tilde{\textbf{m}}}^{2}_{t} \right), Att\left( {\tilde{\textbf{m}}}^{3}_{t} \right) \right]\right)\,,
	\end{aligned}
	\end{equation}
	where $Att$ is the attention unit and $Aggr$ is implemented as convolution layers. Our results presented in Sec.~\ref{part:ablation_cn-cam} demonstrate that our designed module reveals more details.
	
	\subsection{Edge-Aware Loss}
	\label{part:eml}
	Usually, reconstructed high-resolution images produced by VSR methods suffer from jagged edges. To alleviate this problem, we propose an edge-aware loss to produce better refined edges. First, an edge detector is used to extract edge information of ground-truth HR images. Then, the detected edge areas are weighted more in loss calculation, enforcing the network to pay more attention to these areas during learning. 
	
	In this paper, we choose the Laplacian filter as the edge detector. Given the ground-truth $\textbf{I}^{H}_{t}$, the edge map $\textbf{I}^{E}_{t}$ is obtained from the detector and the binary mask value at ${\textbf p}_{t}$ is represented as
	\begin{equation}
	\begin{aligned}
	\textbf{B}_{t} \left( {\textbf p}_{t} \right) &=
	\begin{cases}
	1, & \textbf{I}^{E}_{t} \left( {\textbf p}_{t} \right) \ge \delta \\
	0, & \textbf{I}^{E}_{t} \left( {\textbf p}_{t} \right) < \delta \,,
	\end{cases}
	\end{aligned}
	\end{equation}
	where $\delta$ is a predefined threshold. Suppose the size of a high-resolution image is $H \times W$. The edge mask is also a $H \times W$ map filled with binary values. The areas marked as edges are 1 while others being 0.
	
	During training, we adopt the Charbonnier Loss, which is defined as
	\begin{equation}
	\begin{aligned}
	L = \sqrt{{\left \| \hat{\textbf{I}}^{H}_{t} -  \textbf{I}^{H}_{t}\right \|} ^{2} + \epsilon^{2}} \,,
	\end{aligned}
	\end{equation}
	where $\hat{\textbf{I}}^{H}_{t}$ is the predicted high-resolution result, and $\epsilon$ is a small constant. The final loss is formulated as
	\begin{equation}
	\begin{aligned}
	L_{final} &= L + \lambda {\left \| \textbf{B}_{t} \circ \left(\hat{\textbf{I}}^{H}_{t} - \textbf{I}^{H}_{t} \right) \right \|} \,,
	\end{aligned}
	\end{equation}
	where $\lambda$ is a coefficient to balance the two terms and $\circ$ is element-wise multiplication.
	
	
	\section{Experiments}\label{sec:exp}
	
	\subsection{Dataset and Evaluation Protocol}
	{\bfseries REDS}~\cite{nah2019ntire} is a realistic and dynamic scene dataset published in NTIRE 2019 challenge. There are a total of 30K images extracted from 300 video sequences. The training, validation and test subsets contain 240, 30 and 30 sequences, respectively. Each sequence has equally 100 images with resolution 720 $\times$ 1280. Similar to that of~\cite{wang2019edvr}, we merge the training and validation parts and divide the data into new training (with 266 sequences) and testing (with 4 sequences) datasets. The new testing part contains the $000$, $011$, $015$ and $020$ sequences.
	
	\vspace{0.07in}
	\noindent{\bfseries Vimeo-90K}~\cite{xue2019video} is a large-scale high-quality video dataset designed for various video tasks. It consists of 89,800 video clips which cover a broad range of actions and scenes. The super-resolution subset has 91,701 7-frame sequences with fixed resolution 448 $\times$ 256, among which training and testing splits contain 64,612 and 7,824 sequences respectively.
	
	\vspace{0.07in}
	
	Peak signal-to-noise ratio (PSNR) and structural similarity index (SSIM)~\cite{wang2004image} are used as metrics in our experiments.
	
	\subsection{Implementation Details}\label{key}
	{\bfseries Network Settings\quad} The network takes 5 (or 7) consecutive frames as input. In feature extraction and reconstruction modules, 5 and 40 (20 for 7 frames) residual blocks~\cite{he2016deep} are implemented respectively with channel size 128. In Figure~\ref{fig:tm-cam}, the patch size is 3 and the maximum displacements are set to $\{3, 5, 7\}$ from low to high resolutions. The $K$ value is set to 4. In the cross-scale aggregation module, we define patch size as 1 and fuse information from 4 scales as shown in Figure~\ref{fig:cn-cam}. After reconstruction, both the height and width of images are quadrupled.
	
	\vspace{0.07in}
	\noindent
	{\bfseries Training\quad} We train our network using eight NVIDIA GeForce GTX 1080Ti GPUs with mini-batch size 3 per GPU. The training takes 600K iterations for all datasets. We use Adam as the optimizer and cosine learning rate decay strategy with an initial value $4e-4$. The input images are augmented with random cropping, flipping and rotation. The cropping size is $64 \times 64$ corresponding to output size $256 \times 256$. The rotation is selected as $90^\circ$ or $-90^\circ$. When calculating the edge-aware loss, we set both $\delta$ and $\lambda$ to 0.1.
	
	\vspace{0.07in}
	\noindent
	{\bfseries Testing\quad} In the testing phase, the output is evaluated without boundary cropping.
	
	\subsection{Ablation Study}
	
	To demonstrate the effectiveness of our proposed method, we conduct experiments for each individual design. For convenience, we adopt a lightweight setting in this section. The channel size of network is set to 64 and the reconstruction module contains 10 residual blocks. Meanwhile, the amount of training iterations is reduced to 200K.
	
	\begin{table}[t]
		\caption{Ablation Study of our proposed modules and loss on the REDS testing dataset. `Baseline' is without using the proposed modules and loss. `TM-CAM' represents the temporal multi-correspondence aggregation module. `CN-CAM' means the cross-scale nonlocal-correspondence aggregation module. `EAL' is the proposed edge-aware loss.}
		\begin{center}
			\begin{threeparttable}
				\renewcommand{\tabcolsep}{2.5pt}
				\begin{tabular}{c c c c | c | c}
					\hline
					\multicolumn{4}{c|}{\bf Components} & \multirow{2}*{\bf PSNR(dB)} & \multirow{2}*{\bf SSIM} \\
					Baseline & TM-CAM & CN-CAM & EAL & ~ & ~ \\
					\hline
					$\surd$ & & & & 28.98 & 0.8280 \\
					$\surd$ & $\surd$ & & & 30.13 & 0.8614 \\
					$\surd$ & $\surd$ & $\surd$ & & 30.25 & 0.8641 \\
					$\surd$ & $\surd$ & $\surd$ & $\surd$ & \textbf{30.31} & \textbf{0.8648} \\
					\hline
				\end{tabular}
			\end{threeparttable}
		\end{center}
		\label{tab:ablation}
		\vspace{-0.05in}
	\end{table}

	\begin{table}[t]
		\caption{Results of TM-CAM with different numbers ($K$) of aggregated temporal correspondences on the REDS testing dataset.}
		\begin{center}
			\renewcommand{\tabcolsep}{12pt}
			\begin{tabular}{c | c c}
				\hline
				$K$ & PSNR(dB) & SSIM \\
				\hline
				1 & 30.19 & 0.8624 \\
				2 & 30.24 & 0.8640 \\
				4 & \textbf{30.31} & 0.8648 \\
				6 & 30.30 & \textbf{0.8651} \\
				\hline
			\end{tabular}
		\end{center}\vspace{-0.2in}
		\noindent
		\label{tab:weight_tm-cam}
	\end{table}
	
	\subsubsection{Temporal Multi-Correspondence Aggregation Module}
	\label{part:ablation_tm-cam}
	
	\begin{figure}[t]
		\begin{center}
			\includegraphics[width=0.88\linewidth]{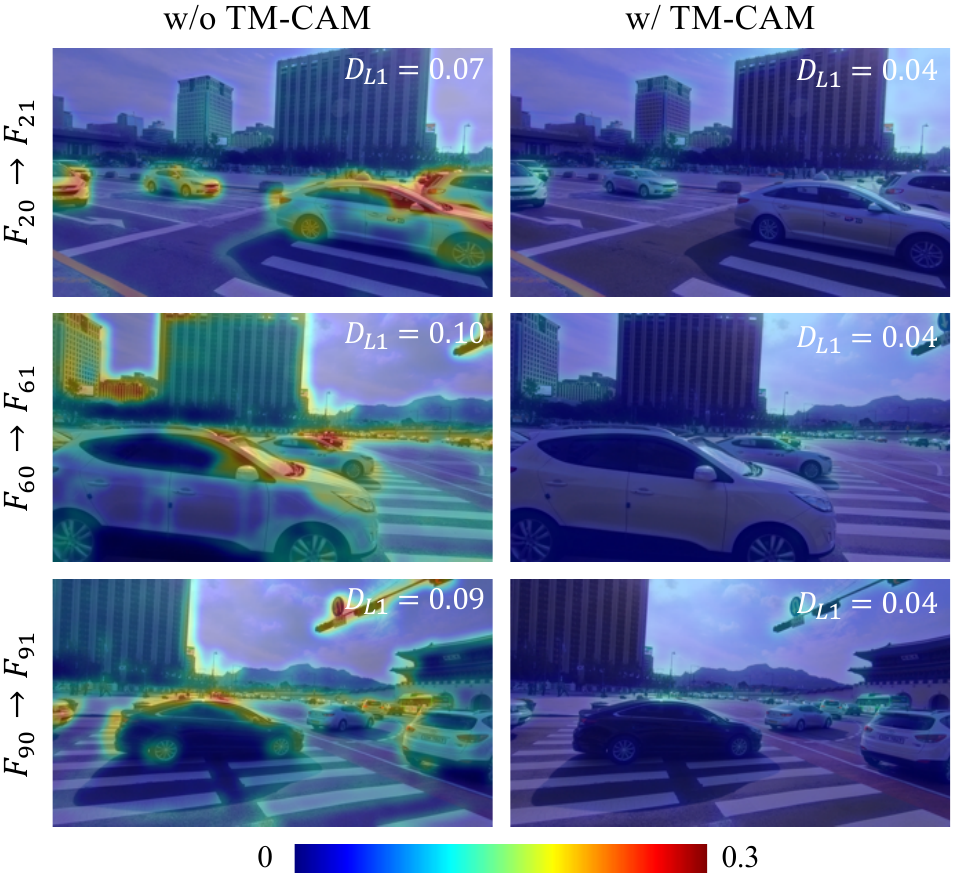}
		\end{center}\vspace{-0.15in}
		\caption{Residual maps between aligned neighboring feature maps and reference feature maps without and with temporal multi-correspondence aggregation module (TM-CAM) on the REDS dataset. Values on the upper right represent the average $L1$ distance between aligned neighboring feature maps and reference feature maps.
			\noindent}
		\label{fig:att_heatmap}
		\vspace{-0.2in}
	\end{figure}
	
	For fair comparison, we first build a baseline without the proposed ideas. As shown in Table~\ref{tab:ablation}, the baseline only yields 28.98dB PSNR and 0.8280 SSIM, a relatively poor result. Our designed alignment module brings about a 1.15dB improvement in terms of PSNR. 
	
	To show effectiveness of TM-CAM in a more intuitive way, we visualize residual maps between aligned neighboring feature maps and reference feature maps in Figure~\ref{fig:att_heatmap}. After aggregation, it is clear that feature maps obtained with the proposed TM-CAM are smoother and cleaner. The mean $L1$ distance between aligned neighboring and reference feature maps are smaller. All these facts manifest the great alignment performance of our method.
	
	We also evaluate how the number of aggregated temporal correspondences affects performance. Table~\ref{tab:weight_tm-cam} shows that the capability of TM-CAM rises at first and drops with the increasing number of correspondences. Compared with taking only one candidate, the four-correspondence setting obtains more than 0.1dB gain on PSNR. It demonstrates that highly correlated correspondences can provide useful complementary details. However, once saturated, it is not necessary to include more correspondences, since weakly correlated correspondences actually bring unwanted noise. We further verify this point by estimating optical flow using a KNN strategy for the MPI Sintel Flow dataset~\cite{butler2012naturalistic}. From Figure~\ref{fig:corr}(b), we find that the four-neighbor setting is also the best choice. Therefore, we set $K$ as 4 in our implementation.
	
	Finally, we verify the performance of pixel-adaptive weights. Based on the experiments, we find that a larger patch in TM-CAM usually gives a better result. It is reasonable since neighboring pixels usually have similar information and are likely to complement each other. Also, structural information is embedded. To balance performance and computing cost, we set the size to 3. When using fixed wieghts (at $K=4$), we obtain the resulting PSNR/SSIM as 30.12dB/0.8614. From Table~\ref{tab:weight_tm-cam}, the proposed pixel-adaptive weighting scheme achieves 30.31dB/0.8648, which is superior to the fixed counterpart by nearly 0.2dB on PSNR, which demonstrates that different aggregating patterns are necessary for consideration of spatial variance.
	
	More experiments of TM-CAM with regard to patch size and maximum displacements are provided in the supplementary file.
	
	\vspace{-0.1in}
	\subsubsection{Cross-Scale Nonlocal-Correspondence Aggregation Module}
	\label{part:ablation_cn-cam}
	In Sec.~\ref{part:ablation_tm-cam}, we already notice that highly correlated temporal correspondences can serve as supplement to motion compensation. To further handle cases with different scales, we proposed a cross-scale nonlocal-correspondence aggregation module (CN-CAM).
	
\begin{figure}[t]
		\begin{center}
			\includegraphics[width=1.0\linewidth]{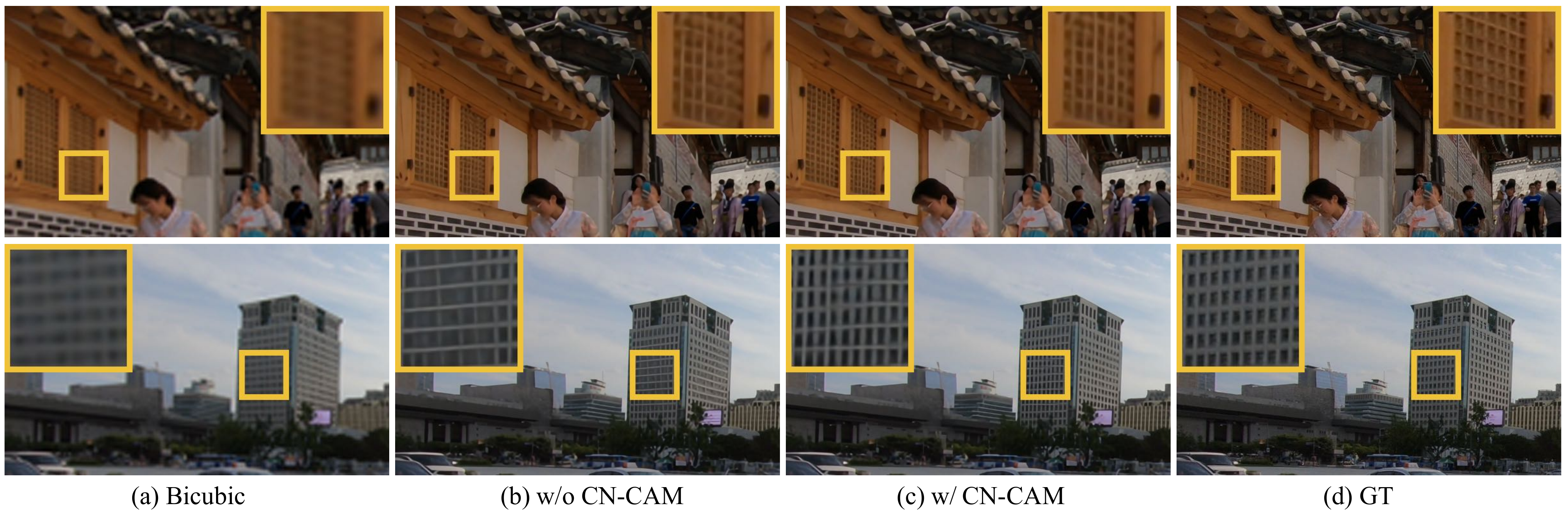}
		\end{center}\vspace{-0.1in}
		\caption{Examples without and with the cross-scale nonlocal-correspondence aggregation module (CN-CAM) on the REDS dataset.}
		\label{fig:cn-cam_ablation}
	\end{figure}

	As listed in Table~\ref{tab:ablation}, CN-CAM improves PSNR by 0.12dB. Besides, Figure~\ref{fig:cn-cam_ablation} reveals that this module enables the network to recover more details when images contain repeated patterns such as windows and buildings within the spatial domain or across scales. All these results show that the proposed CN-CAM method further enhances the quality of reconstructed images.
	
	\begin{figure}[t]
		\begin{center}
			\includegraphics[width=1.0\linewidth]{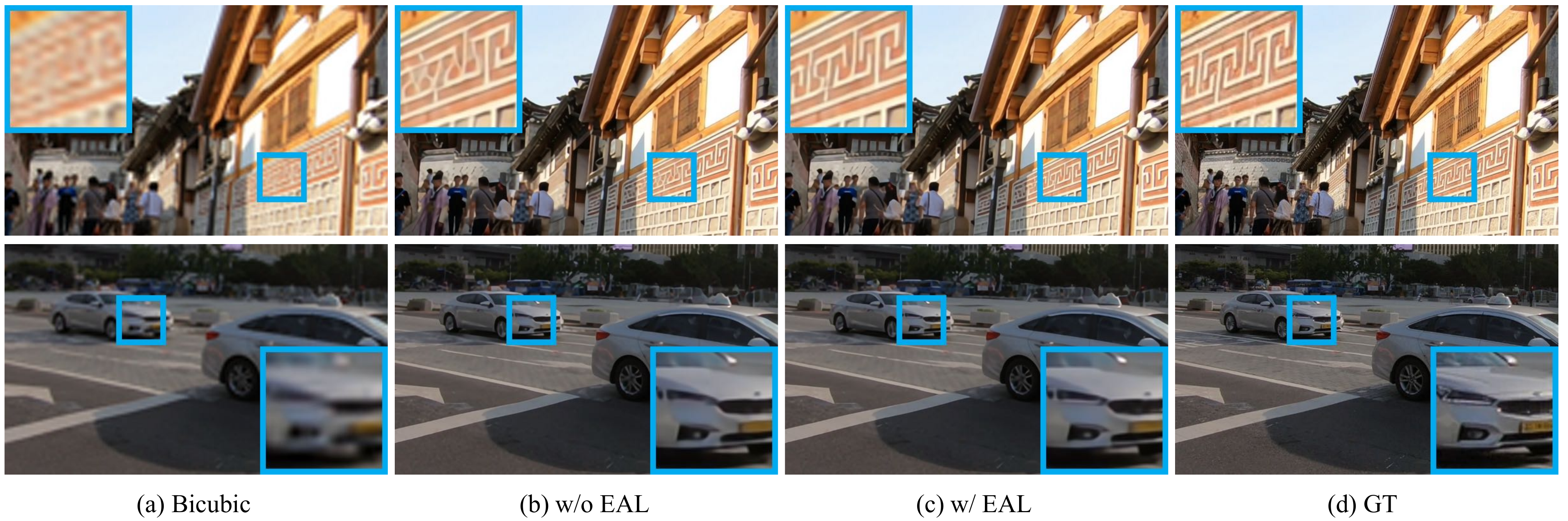}
		\end{center}\vspace{-0.1in}
		\caption{Examples without and with the proposed edge-aware loss (EAL) on the REDS dataset.}
		\label{fig:eal_ablation}
		\vspace{-0.1in}
	\end{figure}
	
	\subsubsection{Edge-Aware Loss}
	In this part, we evaluate the proposed edge-aware loss (EAL). Table~\ref{tab:ablation} lists the statistics. A few visual results in Figure~\ref{fig:eal_ablation} indicate that EAL improves the proposed network further, yielding more refined edges. The textures on the wall and edges of lights are clearer and sharper, which demonstrates the effectiveness of the proposed edge-aware loss.

	\subsection{Comparison with State-of-the-art Methods}
	We compare our proposed multi-correspondence aggregation network (MuCAN) with previous state-of-the-arts including TOFlow~\cite{xue2019video}, DUF~\cite{jo2018deep}, RBPN~\cite{haris2018deep}, and EDVR~\cite{wang2019edvr} on REDS~\cite{nah2019ntire}, Vimeo-90K~\cite{xue2019video} and Vid4~\cite{liu2013bayesian} datasets. The quantitative results in Tables~\ref{tab:res_reds} and \ref{tab:res_vimeo} are extracted from the original publications. Especially, original EDVR~\cite{wang2019edvr} is initialized with a well-trained model. For fairness, we use author-released code to train EDVR without pretraining.
	
\begin{figure}[h]
		\begin{center}
			\includegraphics[width=1.0\linewidth]{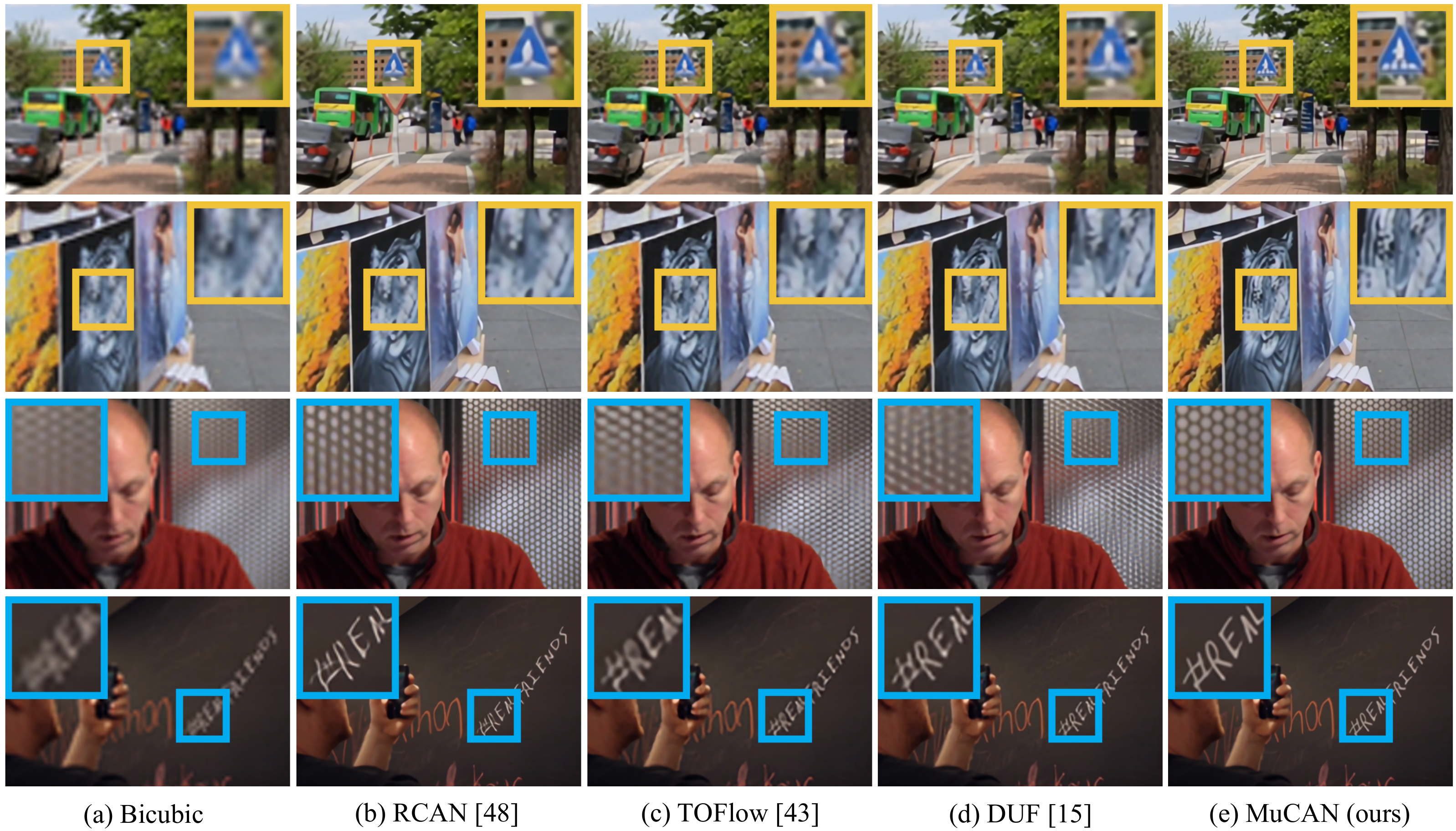}
		\end{center}\vspace{-0.22in}
		\caption{Examples of REDS (top two rows) and Vimeo-90K (bottom two rows) datasets.}
		\label{fig:examples}
	\end{figure}

	On the REDS dataset, results are shown in Table~\ref{tab:res_reds}.  It is clear that our method outperforms all other methods by  at least \textbf{0.17dB}. For Vimeo-90K, the results are reported in Table~\ref{tab:res_vimeo}. Our MuCAN method works better than DUF~\cite{jo2018deep} with nearly \textbf{1.2dB} enhancement on RGB channels. Meanwhile, it brings \textbf{0.25dB} improvement on the Y channel compared with RBPN~\cite{haris2018deep}. All these results verify the effectiveness of our method. Besides, performance on Vid4 is reported in the supplementary file. Several examples are visualized in Figure~\ref{fig:examples}. 
	
	\begin{table}[t]
		\renewcommand\arraystretch{1.3}
		\caption{Comparisons of PSNR(dB)/SSIM results on the REDS dataset for $\times 4$ setting. `*' denotes without pretraining.}\vspace{-0.1in}
		\begin{center}
			\resizebox{\textwidth}{!}{
				\begin{tabular}{c | c | c c c c | c}
					\hline
					Method & Frames & Clip\_000 & Clip\_011 & Clip\_015 & Clip\_020 & Average\\
					\hline
					Bicubic & 1 & 24.55/0.6489 & 26.06/0.7261 & 28.52/0.8034 & 25.41/0.7386 & 26.14/0.7292\\
					RCAN~\cite{zhang2018image} & 1 & 26.17/0.7371 & 29.34/0.8255 & 31.85/0.8881 & 27.74/0.8293 & 28.78/0.8200 \\
					TOFlow~\cite{xue2019video} & 7 & 26.52/0.7540 & 27.80/0.7858 & 30.67/0.8609 & 26.92/0.7953 & 27.98/0.7990 \\
					DUF~\cite{jo2018deep} & 7 & 27.30/0.7937 & 28.38/0.8056 & 31.55/0.8846 & 27.30/0.8164 & 28.63/0.8251 \\
					EDVR*~\cite{wang2019edvr} & 5 & 27.78/0.8156 & 31.60/0.8779 & 33.71/0.9161 & 29.74/0.8809 & 30.71/0.8726 \\
					{\bf{MuCAN}} (Ours) & 5 & {\bf{27.99}}/{\bf{0.8219} }& {\bf{31.84}}/{\bf{0.8801}} & {\bf{33.90}}/{\bf{0.9170}} & {\bf{29.78}}/{\bf{0.8811}} & {\bf{30.88}}/{\bf{0.8750}} \\
					\hline
			\end{tabular}}
		\end{center}
		\label{tab:res_reds}
		\vspace{-0.1in}
	\end{table}
	
	\begin{table}[t]
		\renewcommand\arraystretch{1.1}
		\caption{Comparisons of PSNR(dB)/SSIM results on the Vimeo-90K dataset for $\times 4$ setting. `--' indicates results not available.}
		\small
		\begin{center}
			\renewcommand{\tabcolsep}{4pt}
			\begin{tabular}{c | c | c | c}
				\hline
				Method & Frames & RGB & Y \\
				\hline
				Bicubic & 1 & 29.79 / 0.8483 & 31.32 / 0.8684 \\
				RCAN~\cite{zhang2018image} & 1 & 33.61 / 0.9101 & 35.35 / 0.9251 \\
				DeepSR~\cite{liao2015video} & 7 & 25.55 / 0.8498 & -- \\
				BayesSR~\cite{liu2013bayesian} & 7 & 24.64 / 0.8205 & -- \\
				TOFlow~\cite{xue2019video} & 7 & 33.08 / 0.9054 & 34.83 / 0.9220 \\
				DUF~\cite{jo2018deep} & 7 & 34.33 / 0.9227 & 36.37 / 0.9387 \\
				RBPN~\cite{haris2018deep} & 7 & -- & 37.07 / 0.9435 \\
				{\bf{MuCAN}} (Ours) & 7 & {\bf{35.49}} / {\bf{0.9344}} & {\bf{37.32}} / {\bf{0.9465}} \\
				\hline
			\end{tabular}
		\end{center}
		\label{tab:res_vimeo}
		\vspace{-0.2in}
	\end{table}
	
	\begin{figure}[h]
		\begin{center}
			\includegraphics[width=0.95\linewidth]{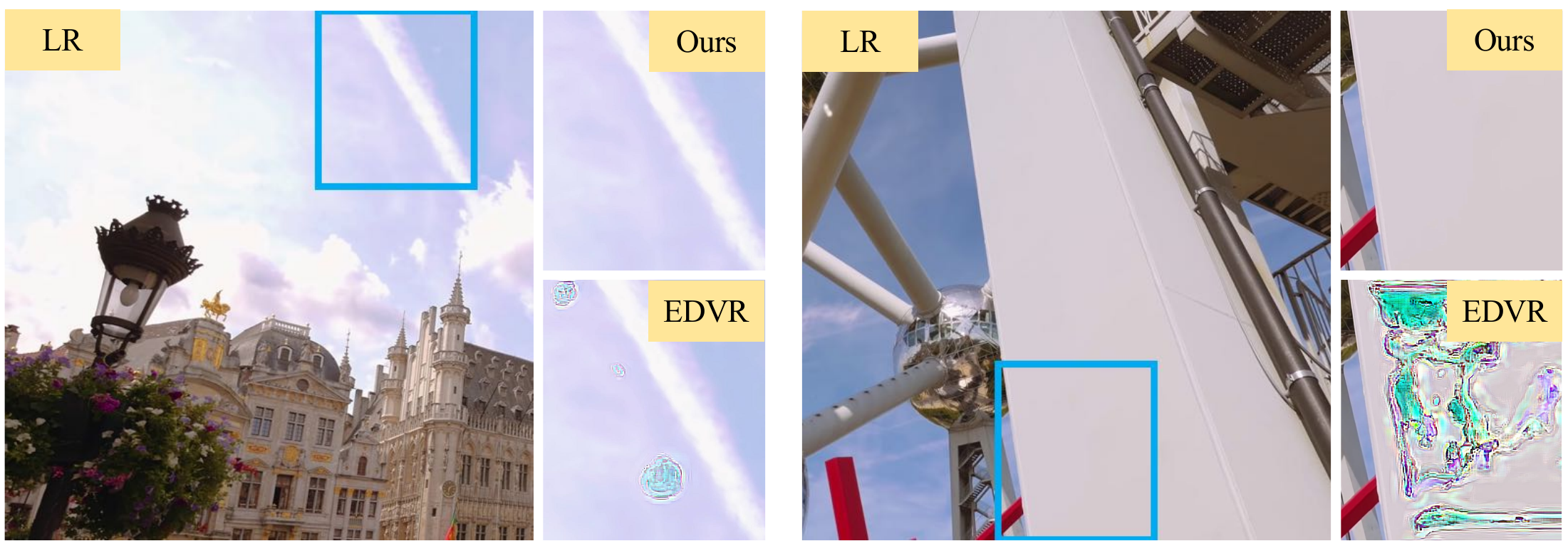}
		\end{center}\vspace{-0.22in}
		\caption{Visualization of video frames in the wild for EDVR~\cite{wang2019edvr} and our MuCAN.}
		\label{fig:general}
	\end{figure}
	
	\subsection{Generalization Analysis}
	\label{part:generalization}
	
	To evaluate the generality of our method, we apply our model trained on the REDS dataset to test video frames in the wild. In addition, we test EDVR with the author-released model\footnote{https://github.com/xinntao/EDVR} on the REDS dataset. Some visual results are shown in Figure~\ref{fig:general}. We remark that EDVR may generate visual artifacts in some cases due to the variance of data distributions between training and testing. In contrast, our MuCAN performs well in the real world setting and shows its decent generality.
	
	
	\section{Conclusion}
	
	In this paper, we have proposed a novel multi-correspondence aggregation network (MuCAN) for the video super-resolution task. We showed that the proposed temporal multi-correspondence aggregation module (TM-CAM) takes advantage of highly correlated patches to achieve high-quality alignment-based frame recovery. Additionally, we verified that the cross-scale nonlocal-correspondence aggregation module (CN-CAM) utilizes multi-scale information and further boosts the performance of our network. Also, the edge-aware loss enforces the network to obtain more refined edges on the high-resolution output. Extensive experiments have manifested the effectiveness and generality of our proposed method.
	
	\clearpage
	%
	%
	\bibliographystyle{splncs04}
	\bibliography{egbib}
\end{document}